\documentclass[letterpaper, 10pt, conference]{ieeeconf}
\IEEEoverridecommandlockouts
\usepackage{graphicx}
\usepackage{float}
\usepackage{url}
\usepackage{cite}
\usepackage{amsmath,amssymb,amsfonts}
\usepackage{algorithmicx}
\usepackage{graphicx}
\usepackage{textcomp}
\usepackage{algorithm}
\usepackage[noend]{algpseudocode}

\def\is#1{\hbox{\scriptsize\it #1\/}}
\def\ii#1{\hbox{\it #1\/}}

\title{\LARGE \bf Object Placement on Cluttered Surfaces:\\ A Nested Local Search Approach}

\author{
    Abdul Rahman Dabbour \and Esra Erdem \and Volkan Patoglu
    \thanks{This work was partially supported by Sabanc{\i} University.}
    \thanks{
        A. R. Dabbour, E. Erdem and V. Patoglu are with the Faculty of Engineering and Natural Sciences, Sabanc{\i} University, \.Istanbul, Turkey.
        {\tt {\scriptsize \{dabbour,vpatoglu,esraerdem\}@sabanciuniv.edu}}
    }
}

\begin{document}

\maketitle
\thispagestyle{empty}
\pagestyle{empty}

\begin{abstract}
For planning rearrangements of objects in a clutter, it is required to know the goal configuration of the objects. However, in real life scenarios, this information is not available most of the time. We introduce a novel method that computes a collision-free placement of objects on a cluttered surface,  while
minimizing the total number and amount of displacements of the existing moveable objects. Our method applies nested local searches that perform multi-objective optimizations guided by heuristics. Experimental evaluations demonstrate high computational efficiency and  success rate  of our method, as well as good quality of solutions.
\end{abstract}

\section{INTRODUCTION}

For useful integration of robotic systems into everyday life, they must be capable of performing high-complexity real-life tasks efficiently. For instance, typical human environments, such as table tops, kitchen shelves, or office desks, are usually cluttered; and manipulating the environment to deal with such clutter is integral to performing everyday chores in social environments, whether that means rearranging objects upon a surface or across multiple surfaces.

Geometric rearrangement of multiple movable objects on a surface is a difficult problem, because it requires the manipulation of existing objects on the surface, as well as the placement of new objects to be put on the surface. To solve such a problem, in general, task planning is required to decide for the order of manipulation actions (e.g., when to pick, place, move objects), and feasibility checks are required to check the execution of each manipulation action against geometric/kinematic constraints (e.g., to avoid collisions). These requirements usually lead to hybrid planning solutions that combine high-level task planning and low-level feasibility checks.

To solve such a planning problem for the rearrangement of objects in a clutter, one needs to know the goal configuration (i.e., how the objects are arranged on the surface at the end of the plan). However, in real life scenarios, this information is not available most of the time. For that reason, it is essential to determine a geometrically feasible goal configuration of objects on the surface before planning for rearrangements.

With this motivation, we study the object placement problem: given a surface cluttered with (unmoveable) obstacles and a set of existing moveable objects on it, and a  set of new objects to be placed on the surface, the goal is to find a collision-free placement of all objects on the surface while minimizing the total number and amount of displacements of the existing moveable objects.

We introduce an efficient algorithm based on nested local searches. The innermost search aims to minimize the total penetration depth of objects utilizing a potential field method over a physics-based engine, the intermediate local search aims to minimize the number of collisions by allowing re-placements of objects, and the outermost local search further tries to minimize the number of displacements of the movable objects on the surface with respect to their initial configurations. The intermediate local search relies on a grid-based heuristics to find more plausible object configurations, while the outermost local search relies on a heuristic that gradually relaxes the constraints imposed on object movements.

\section{RELATED WORK}

{\em Related work in robotics} Rearrangement of multiple movable objects, a challenging  problem that involves planning, manipulation and geometric reasoning, has received much attention in robotics. In particular,  planning for geometric rearrangement with multiple movable objects and its variations, such as navigation among movable obstacles~\cite{stilman2005navigation,stilman2008planning}, have been studied using various approaches. Since even a simplified variant the rearrangement problem with only one movable obstacle has been proved to be NP-hard~\cite{wilfong1991motion,demaine2003pushing}, most studies introduce several important restrictions to the problem, like monotonicity of plans~\cite{okada2004environment,stilman2007manipulation,dogar2012planning,barry2013manipulation,cosgun2011push}, where each object can be moved at most once. Recent work have focused on generating non-monotonic plans~\cite{havur2014geometric,krontiris2014rearranging,krontiris2015dealing,krontiris2016efficiently,han2017high,kang2018automated}. However, in most of these studies~\cite{okada2004environment,stilman2007manipulation,dogar2012planning,barry2013manipulation,krontiris2014rearranging,krontiris2015dealing,krontiris2016efficiently,han2017high}, it is assumed that the goal configuration is known. Finding suitable arrangements for objects on a cluttered surface has received relatively less attention.

Cosgun et al.~\cite{cosgun2011push} propose an algorithm that searches for a suitable placement for a single object on a cluttered surface by discretizing the possible orientations of the object, convolving object pixels with the ones on the table, and identifying candidate regions for the object placement that result in minimal penetration with other objects. A placement is then produced by sampling these regions; however, this placement may not be collision-free. Then, they plan for a sequence of linear push actions to rearrange the clutter  and  clear space for the new object such that this placement becomes collision-free. Note that there are several limitations in this approach; multiple new objects are not considered, the surface and possible object orientations are discretized, and the final configuration is not necessarily collision-free.

Yu et al.~\cite{yu2011make} aim to find sensible placements for furniture by initially generating a random arrangement, then rearranging it to minimize a cost function that measures the difference between the current arrangement and several positive examples provided by the user. Kang et al.~\cite{kang2018automated} also follow on this idea, and modify the algorithm so it becomes more suitable for robotic applications. Neither study considers heavily cluttered scenes or utilizes high resolution collision checks. In~\cite{kang2018automated}, the task is to rearrange objects currently available in the scene to achieve  a more tidy arrangement; no new objects are added and there exists no constraints that force a certain set of objects to be on certain surfaces. Furthermore, in these studies, even the initial state is a feasible (collision-free) configuration, and the goal is to improve it in terms of a measure of tidiness.

Jiang et al.~\cite{jiang2012learning,jiang2012humanlearning,jiang2013hallucinating} extract object-to-object and object-to-human features from databases of 3D environments and learn semantic/geometric preferences for object surface pairs. Then, they discretize the surfaces' point cloud into placing areas by random sampling and solve an maximum matching problem to assign each object's pose to a suitable placing area. This approach only considers placements to a predetermined set of discrete configurations and does not address the more challenging continuous version of the problem.

In our previous work~\cite{havur2014geometric}, we have proposed an object placement algorithm based on a local search guided with heuristics and random restarts. This work significantly extends our earlier study by introducing an innermost potential field, as well as two nested local searches  wrapped around this basic search algorithm, to improve upon the efficiency and quality of solutions, as well as the success rate. Our results indicate orders of magnitude difference in terms of CPU time and success rate in cluttered scenarios.

{\em Related work in other areas} A closely related problem to object placement, studied in computer graphics and operations research, is the packing problem (also known as the knapsack problem), where the goal is to place as many objects as possible in a non-overlapping configuration within a given empty container. The packing problem is NP-hard~\cite{Chazelle1989}. It has been widely studied in 2D (cf. the survey~\cite{DYCKHOFF1990}). It has been also studied in 3D under various conditions/restrictions~\cite{Liu2015,ROMANOVA2018,Ma2018} (e.g., packing a set of polyhedrons into a fixed size polyhedron without considering rotations~\cite{EGEBLAD2009}, orthogonal packing of tetris-like items into rectangular bins~\cite{martello2000,Fasano2013}).

However, the object placement problem is quite different from these 3D packing problems. First of all, since object placement problem is motivated by the geometric rearrangement of objects on a cluttered surface, the surface does not have to be empty and contains movable objects. The packing problem, on the other hand, assumes that the fixed size container is empty. Also the objective function for the object placement problem is different: the goal is to find a collision-free configuration of all objects, so as to minimize the total number and amount of displacements of the existing objects on the surface. The packing problem, on the other hand, aims to find a collision-free configuration of some objects, so as to maximize the coverage rate (i.e., the total volume of the objects packed in the container). Along these lines, the packing problem and the placement problem are different computational problems with different optimization goals. The methods to attack these problems are significantly different from each other and do not allow for direct comparisons.

\begin{figure*}[t]
    \centering
    \resizebox{2.05\columnwidth}{!}{\includegraphics{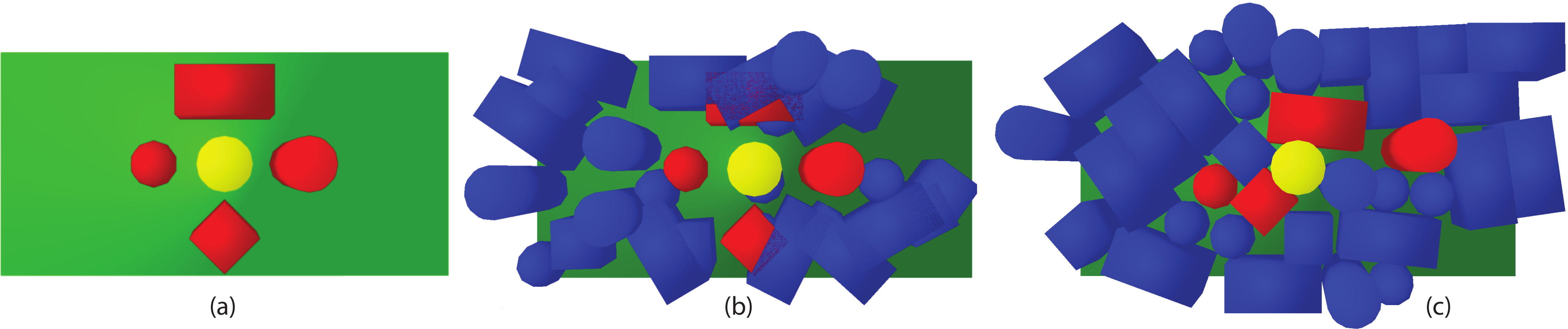}}
    \vspace{-1.5\baselineskip}
    \caption{(a) A surface with an obstacle and four movable objects. (b) A random placement of new objects on the surface. (c) A collision-free placement of all objects.}
    \vspace{.25\baselineskip}
    \label{fig:potential_field}
\end{figure*}

\section{PROBLEM DEFINITION}

The \textit{object placement problem} is defined by

\begin{itemize}
    \item a flat surface $S$ and its geometric model $g(S)$ that details its size and shape,
    \item the set $O_O$ of non-movable objects (obstacles) on the surface $S$ and the set $g(O_O)$ of their geometric models,
    \item the set $O_C$ of movable objects (clutter) on the surface $S$ and the set $g(O_C)$ of their geometric models,
    \item the set $O_N$ of new objects to be placed on the surface $S$ and the set $g(O_N)$ of their geometric models,
    \item a set $W$ of continuous placement constraints on objects (e.g., a monitor may be forced to be in the corner of the table), and
    \item an initial collision-free configuration $C_I$ of all objects in $O_C\cup O_O$ on the surface $S$ relative to $g(S)$.
\end{itemize}

All objects are assumed to be rigid bodies.

Before we define a solution for an object placement problem, let us introduce the following definitions and notations for a configuration $C_x$ of all objects $O_N \cup O_C$ on the surface $S$ relative to $g(S)$.
\begin{itemize}
\item
The number of displacements $\ii{\#move}(C_x)$ in $C_x$ is the number of objects in $O_C$ whose configurations are different in $C_x$ from their original configurations in $C_I$.
\item
The amount of displacement $\ii{dist}_o(C_x)$ of an object $o\in O_C$ in $C_x$ is the distance between the centroid of $o$ in $C_{I}$ and the centroid of $o$ in $C_x$. The total amount of displacements of objects in $O_C$ is $\sum_{o\in O_C} \ii{dist}_o(C_x)$.
\item
The amount of change $\ii{arc}_o(C_x)$ in the orientation of an object $o\in O_C$ in $C_x$ is the arc length traced by the most distal point on the object, due to the change of configuration of $o$ from $C_{I}$ to $C_x$. The total amount of change in orientations of objects in $O_C$ is $\sum_{o\in O_C} \ii{arc}_o(C_x)$.
\item
The total amount of change $\ii{cost}_d(C_x)$ in configurations of objects in $O_C$ with respect to $C_I$ is the sum of the total amount of displacements of objects in $O_C$ (i.e., $\sum_{o\in O_C} \ii{dist}_o(C_x)$) and the total amount of change in orientations of objects in $O_C$ (i.e., $\sum_{o\in O_C} \ii{arc}_o(C_x)$).
\end{itemize}

A solution for an object placement problem $\langle g(S)$, $O_O$, $g(O_O)$, $O_C$, $g(O_C)$, $O_N$, $g(O_N)$, $C_I$, $W\rangle$ is a collision-free final configuration ${C}_F$ of all objects $O_N \cup O_C$ on the surface $S$ relative to $g(S)$, with a minimal value of $\langle \ii{\#move},\ii{cost}_d\rangle$ relative to lexicographic ordering (i.e., there does not exist a collision-free configuration $C_x$ such that $\ii{\#move}(C_x){<} \ii{\#move}(C_F)$, or  $\ii{\#move}(C_x){=}\ii{\#move}(C_F)$ and $\ii{cost}_d(C_x){<} \ii{cost}_d(C_F)$.

Figure~\ref{fig:potential_field} presents a sample problem instance: initially in Figure~\ref{fig:potential_field}(a), a cylindrical obstacle (yellow), and four geometrically different movable objects (red) are placed on a surface (green); the goal is to find a collision-free configuration of these objects and some more new objects (blue), like in Figure~\ref{fig:potential_field}(c).

\section{METHOD}

We propose a nested local search algorithm (Algorithm~\ref{alg:outermost}) to compute a solution for an object placement problem. Intuitively; the innermost search applies potential field method over a physics-based engine, with the goal of minimizing the total penetration depth of objects, the intermediate local search further aims to minimize the number of collisions by allowing re-placements of objects, and the outermost local search further tries to minimize the number and amount of displacements of movable objects with respect to their initial configurations.

\begin{algorithm}
	\caption{$\textsc{outermostSearch}$}{
	\label{alg:outermost}
	\textbf{Input:} Object placement problem $P=\langle g(S)$, $O_O$, $g(O_O)$, $O_C$, $g(O_C)$, $O_N$, $g(O_N)$, $C_I$, $W\rangle$. \\
	\textbf{Output:} A configuration of all objects $O_N \cup O_C$ on the surface $S$.
\begin{algorithmic}[1]
		\Statex // $V$: a set of placement constraints for every object $o\in O_C$, specifying $\ii{radius}_o$ of the balls where $o$ can be placed in
		\Statex // $\ii{cost}_r$: cost function characterizes the total amount of changes of object poses with respect to $C_I$

        \State $C_{\is{current}} \gets$ a configuration of all objects $O_O\cup O_C \cup O_N$ on the surface $S$ obtained from $C_I$ by randomly placing the new objects $O_N$ on the surface $S$;
        \State $V\gets$ for every $o\in  O_C$, $\ii{radius}_o{=}0$

        \Statex // Call the intermediate local search to reduce number of collisions and check if it also decreases the total pose changes
		\State $C_{\is{next}} \gets \textsc{intermediateSearch}(P,V,C_{\is{current}})$;
    	\If{$\ii{cost}_r(C_{\is{next}}) \prec \ii{cost}_r(C_{\is{current}})$}
          \State $C_{\is{current}} {=} C_{\is{next}}$;
        \EndIf
        \Statex // Relax the placement constraints and call the intermediate local search until no better configuration can be found
        \State $C_{\is{best}} {=} C_{\is{current}}$;
        \Loop
          \For{$o\in O_C$ where $\ii{radius}_o{<}1$}
              \State $\ii{radius}_o \gets$ increase the radius slightly;
              \State $V_o\gets$ update $V$ with $\ii{radius}_o$;
              \State $C_{o} \gets\textsc{intermediateSearch}(P,V_o,C_{\is{current}})$;
              \If{$\ii{cost}_r(C_{o}) \prec \ii{cost}_r(C_{\is{best}})$}
                 \State $C_{\is{best}}{=}C_{o}$;
              \EndIf
          \EndFor
      \If{$\ii{cost}_c(C_{\is{current}}) \preceq \ii{cost}_c(C_{\is{best}})$}
        \State return $C_{\is{current}}$;
      \Else
        \State $C_{\is{current}} {=} C_{\is{best}}$;
      \EndIf
        \EndLoop
	\end{algorithmic}}
\end{algorithm}
\setlength{\textfloatsep}{\baselineskip}
\setlength{\floatsep}{\baselineskip}

\begin{algorithm}
	\caption{$\textsc{intermediateSearch}$} {
	\label{alg:intermediate}
	\textbf{Input:} Object placement problem $P=\langle g(S)$, $O_O$, $g(O_O)$, $O_C$, $g(O_C)$, $O_N$, $g(O_N)$, $C_I$, $W\rangle$; placement constraints $V$; and, if provided, a configuration $C_{\is{current}}$ of all objects $O_O\cup O_C \cup O_N$ on the surface $S$ ($C_{\is{current}}$ is obtained from $C_I$ in \textsc{outermostSearch})\\
	\textbf{Output:} A configuration of all objects $O_N \cup O_C$ on the surface $S$.
	\begin{algorithmic}[1]
	\Statex // $H$: a set of free cells, suggested re-placements of objects $o\in O_C$ in collision
	\Statex // $\ii{cost}_c$: cost function characterizes the total number of collisions ($\ii{\#col}$) and the total amount of penetration depth of pairs of objects in collisions ($\ii{cost}_p$) considering the placement constraints $V$

    \State $C_{\is{current}} \gets$ if not provided, generate a configuration of all objects $O_O\cup O_C \cup O_N$ on the surface $S$ obtained from $C_I$ by randomly placing the new objects $O_N$ on the surface $S$

    \Statex // Call the dynamic simulator to reduce the total penetration depth, and check if it also decreases the number of collisions
	\State $C_{\is{next}} \gets \textsc{innermostSearch}(P, V, C_{\is{current}})$;
	\If{$\ii{cost}_c(C_{\is{next}}) \prec \ii{cost}_c(C_{\is{current}})$}
      \State $C_{\is{current}} {=} C_{\is{next}}$;
    \EndIf

    \Statex // Re-place objects in collisions and call the dynamic simulator until no better configuration can be found
    \State $C_{\is{best}} {=} C_{\is{current}}$;
    \Loop
      \State $H\gets$ refine discretization, identify free cells
      \For{$o\in O_C \cup O_N$ in collision}
         \For{every free cell $e$ in $H$}
           \State $C_o \gets$ re-place $o$ in $C_{\is{current}}$ in $e$
           \State $C_x \gets\textsc{innermostSearch}(P,C_o)$;
           \If{$\ii{cost}_c(C_x) \prec \ii{cost}_c(C_{\is{best}})$}
             \State $C_{\is{best}}{=}C_x$;
           \EndIf
         \EndFor
      \EndFor
      \If{$\ii{cost}_c(C_{\is{current}}) \preceq \ii{cost}_c(C_{\is{best}})$}
        \State return $C_{\is{current}}$;
      \Else
        \State $C_{\is{current}} {=} C_{\is{best}}$;
      \EndIf
    \EndLoop
\end{algorithmic}}
\end{algorithm}

\subsection{Penetration Minimization}

The innermost search relies on  artificial potential fields~\cite{Khatib1086} defined for each object (including the obstacles and boundaries of the table) and a dynamic simulator to calculate a configuration in static equilibrium. In particular, trajectories of objects under the action of interaction forces due to potential fields are simulated according to the governing equations of motion, that are, effectively, a solution to a dynamic optimization problem over the system's Lagrangian. The algorithm starts with a configuration $C_{\is{current}}$ obtained from $C_I$ by randomly placing the new objects $O_N$ on the surface $S$, with zero velocities. It searches over configurations of all objects $O_O\cup O_C \cup O_N$ on the surface $S$, with the goal of minimizing the objective optimization function $\ii{cost}_p$ defined as the sum of maximum penetration depth~\cite{Patoglu2005a} of pairs of objects in collisions.  The physics engine returns a configuration $C_{\is{next}}$ of all objects, minimizing $\ii{cost}_p(C_{\is{next}})$.

\begin{figure}[t]
    \centering
    \resizebox{\columnwidth}{!}{\includegraphics{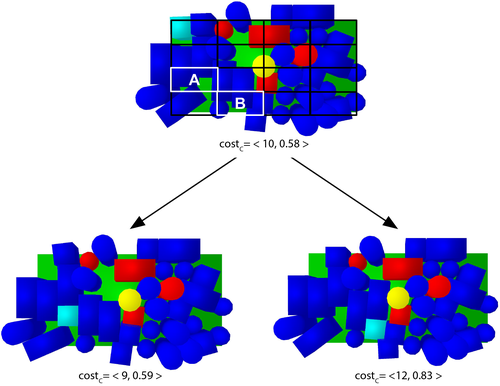}}
    \vspace{-1.5\baselineskip}
    \caption{Intermediate local search algorithm}
    \label{fig:relocation_example}
\end{figure}

For instance, consider the initial configuration $C_I$ of obstacles (yellow) and movable objects (red) shown in Figure~\ref{fig:potential_field}(a). A configuration $C_{\is{current}}$ of objects obtained by randomly placing all the new objects (blue) on the surface can be seen in Figure~\ref{fig:potential_field}(b). Note that there are many collisions in $C_{\is{current}}$, and objects penetrate with each other. By applying this innermost search, the configuration shown in Figure~\ref{fig:potential_field}(c) may be generated, where the total penetration is zero, since there is no collisions.

The only action that the physics engine can perform (due to the use of the potential field method) is to push objects in collision outside of each other; so, for instance, it cannot swap locations of two objects. This may lead to local minima where the objects can no longer be pushed, and there may be still some collisions in $C_{\is{next}}$. This motivates us towards the intermediate local search algorithm that utilizes some heuristics to re-place all the objects in collision.

\begin{figure*}[t]
    \centering
    \resizebox{1.2\columnwidth}{!}{\includegraphics{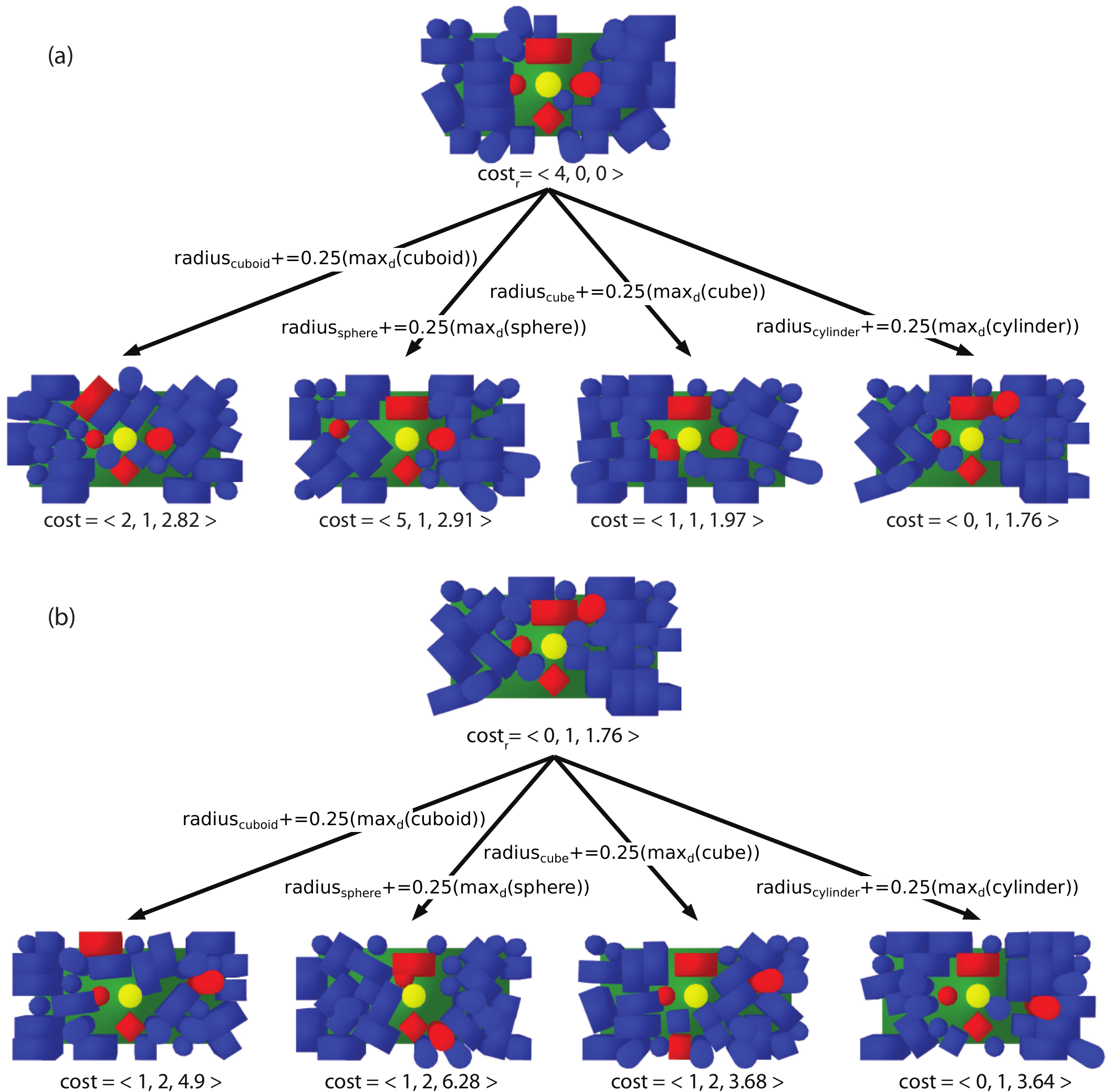}}
    \vspace{-0.5\baselineskip}
    \caption{Outermost local search algorithm}
    \vspace{-.5\baselineskip}
    \label{fig:outer_search}
\end{figure*}

\subsection{Collision Minimization}

The intermediate local search algorithm (Algorithm~\ref{alg:intermediate}) utilizes heuristically-guided re-placements to avoid local minima. Intuitively, for each object $o$ in collision, the heuristic suggests (i) dividing the surface $S$ into cells and (ii) re-initiating the placement of the object $o$  into the center one of the cells that is unoccupied (i.e., has no other object centroids); the heuristics is used to guide generation of new configurations. For (i), the heuristic imposes a grid on the surface $S$. If a grid cell has no object centroids in it, it is marked as free. If no free cell exists, a new refined grid is imposed on $S$ with smaller but more number cells. This refinement process continues until when there is at least one free cell in the imposed grid.

The intermediate local search algorithm also starts with a configuration $C_{\is{current}}$ of all objects $O_O\cup O_C \cup O_N$ on the surface $S$ obtained from $C_I$ by randomly placing the new objects $O_N$ on the surface $S$. It calls the innermost search algorithm described above to find a better configuration, but starting with the configurations obtained from $C_{\is{current}}$ as suggested by the heuristics. The goal is to minimize the objective optimization function $\ii{cost}_c$ defined as a tuple $\langle \ii{\#col}, \ii{cost}_p\rangle$, where $\ii{\#col}$ is the total number of collisions and $\ii{cost}_p$ is the total amount of penetration depth of pairs of objects in collisions. Here, lexicographic ordering is used to find the minimum of two tuples: $\langle a,b\rangle \prec \langle a',b'\rangle$ if either $(a \prec a')$ or  ($a=a'$ and $b\prec b'$). In this way, priority is given to $\ii{\#col}$ and then to $\ii{cost}_p$: among multiple configurations of all objects with the same minimum number of collisions, the configuration $C_{\is{next}}$ with the least cumulative penetration depth is returned.

An example is presented in Figure~\ref{fig:relocation_example} to illustrate the usefulness of the heuristics in  the intermediate local search algorithm. The search starts with a configuration $C_{\is{current}}$, where ${\ii{\#col}(C_{\is{current}})=24}$ and  ${\ii{cost}_p(C_{\is{current}})=1.75}$. Note that the innermost search alone cannot find a better configuration with less value of $\ii{cost}_p$: the physics engine gets stuck, as it can no longer push objects on the right half of the surface. First, the heuristic is utilized to re-place the cyan-colored object, which is in collision with some other object, in $C_{\is{current}}$. For that, a grid is imposed over the surface with two free cells, labelled A and B, that do not contain any object centroids. Then, two new configurations, $C_{\is{A}}$ and $C_{\is{B}}$, are obtained from $C_{\is{current}}$ by randomly re-placing the cyan-colored object in A and in B, respectively. Here, ${\ii{cost}_c(C_{\is{A}}) = \langle 19, 1.43 \rangle}$ and ${\ii{cost}_c(C_{\is{B}}) = \langle 20, 1.37 \rangle}$. At this point, the intermediate local search algorithm calls the innermost search algorithm for each of these two configurations.

The intermediate local search algorithm with heuristically-guided re-placements is useful for minimizing the number of collisions on a surface, but the objects in $O_C$ may end up being displaced and rotated too much with respect to their original configurations in $C_I$.  This is undesirable from the perspective of rearrangement planning, because it will require more number of manipulation actions to rearrange such objects. This motivates us towards the outermost local search algorithm, which utilizes some constraints on the placements of objects to limit their movements.

\subsection{Rearrangement Minimization}

The outermost local search algorithm (Algorithm~\ref{alg:outermost}) utilizes placement constraints to minimize displacements.  Intuitively, the amount of displacement of a movable object $o\in O_C$ is constrained to a ball whose centroid is the object's centroid in the given initial configuration in $C_I$. Initially, $\ii{radius}_o$ is set to $0$; if the outermost local search algorithm cannot find a better configuration under these constraints, then $\ii{radius}_o$ is increased slightly.

The outermost local search algorithm starts with a configuration $C_{\is{current}}$ of all objects $O_O\cup O_C \cup O_N$ on the surface $S$ obtained from $C_I$ by randomly placing the new objects $O_N$ on the surface $S$. It calls the intermediate local search algorithm described above to find a better configuration, but with respect to the given set $V$ of placement constraints. The goal is to minimize the objective optimization function $\ii{cost}_r$ defined as a triple $\langle \ii{\#col}, \ii{\#move}, \ii{cost}_d\rangle$, where $\ii{\#col}$ is the total number of collisions, $\ii{\#move}$ is the total number of moves of objects in $O_C$ from their original places, and $\ii{cost}_d$ is the total amount of change in configurations of objects in $O_C$ with respect to $C_I$. Here, $\ii{cost}_d$ for a configuration $C_x$ is computed as the sum of the total amount of displacements of objects in $O_C$ (i.e., $\sum_{o\in O_C} \ii{dist}_o(C_X)$, where $\ii{dist}_o(C_X)$ is the distance between the centroid of $o$ in $C_{I}$ and the centroid of $o$ in $C_x$) and the total amount of change in orientations  of objects in $O_C$ (i.e., $\sum_{o\in O_C} \ii{arc}_o(C_X)$, where $\ii{arc}_o(C_X)$ is the arc length due to the change of configuration of $o$ from $C_{I}$ to $C_x$). The outermost local search algorithm also uses lexicographic ordering to find the minimum of two triples. In this way, priority is given first to $\ii{\#col}$, then to \ii{\#move}, and then to $\ii{cost}_d$.

An example is presented in Figure~\ref{fig:outer_search} to illustrate the usefulness of the placement constraints in the outermost local search algorithm.
The search starts with a configuration $C_{\is{current}}$, where ${\ii{\#col}(C_{\is{current}})=4}$ and  $\ii{\#move}(C_{\is{current}})=\ii{cost}_d(C_{\is{current}})=0$. Initially, the radius of the placement balls for every object that is initially on the table is zero. With these placements constraints, the intermediate local search cannot find a better configuration to optimize $\ii{cost}_r$. Then, for each object, the outermost local search algorithm relaxes its placement constraints by increasing the radius of its placement ball by a certain percentage of $max_d(o)$, where $max_d(o)$ is the distance from the centroid of the object to the furthest point of the surface, then calls the intermediate local search again. With such relaxed constraints, the intermediate local search algorithm returns better configurations with less costs. For example, when it is allowed to place the cuboid object within a small circle around its initial configuration, the intermediate local search algorithm returns a configuration $C_{\is{cuboid}}$ with ${\ii{cost}_r(C_{\is{cuboid}}) = \langle 2, 1, 2.82 \rangle}$; whereas, when it is allowed to displace the cylindrical object by a small amount, it returns a configuration $C_{\is{cyl}}$ with ${\ii{cost}_r(C_{\is{cyl}}) = \langle 0, 1, 1.76 \rangle}$ (Figure~\ref{fig:outer_search}(a)).  The outermost local search algorithm continues search from $C_{\is{cyl}}$, with even more relaxed placement constraints, but cannot find a better configuration with less cost; so the outermost search stops (Figure~\ref{fig:outer_search}(b)).

\section{EXPERIMENTAL EVALUATION}
\label{sec:exp}

\begin{figure*}[t]
    \centering
        \resizebox{2.05\columnwidth}{!}{\includegraphics{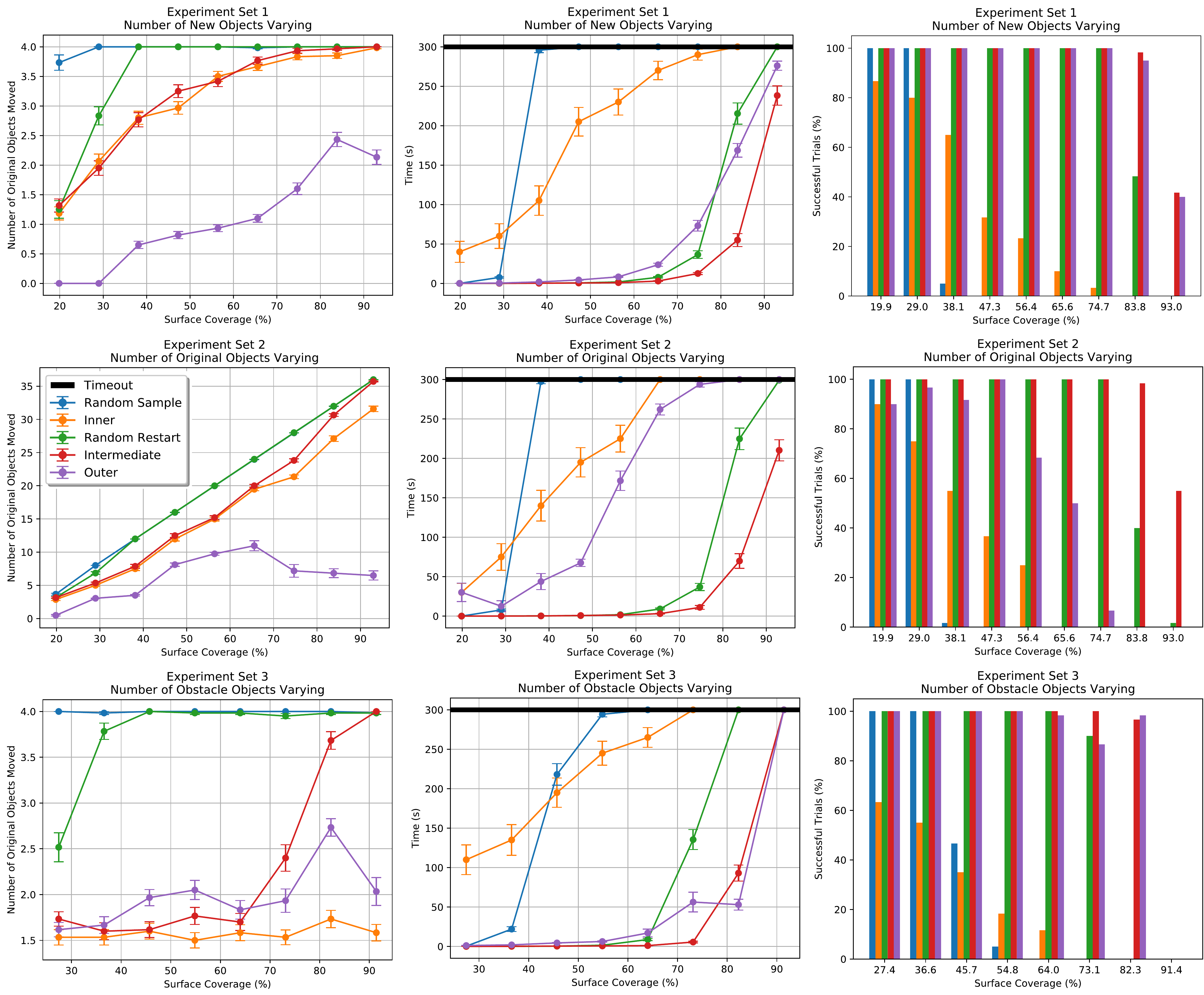}}
    \vspace{-1.5\baselineskip}
    \caption{Plots summarize the experimental results, where each instance of each experiment was run 60 times for averaging of the results. Rows correspond to different experiments, while columns present various metrics. In particular, CPU time, solution quality, and success rate metrics are presented in the columns.  Experiment~1 varies the number of new objects, Experiment~2 varies the number of obstacles, and Experiment~3 varies the number of movable objects that are initially on the table.}
    \label{fig:exp_plots}
\end{figure*}

We have conducted three sets of experiments to evaluate and compare performance of each search level. All simulations were executed on workstation with an Intel Xeon W-2155 CPU running at 3.30~GHz using a single thread and 32~GB RAM. All algorithms were implemented in Python, with Bullet~\cite{coumans2010bullet} as the back-end physics engine. Each instance of each experiment was run 60 times to allow for averaging of the results. A timeout of 300~s per trial was imposed.

Experiment~1 started with the initial configuration depicted in Figure~1(a), where 4  objects and an obstacle were on a surface. The number of new objects to be added to the table was the control variable in this condition. The number of new objects was increased from 4 to 36, such that the total footprint area covered by all objects and the obstacle were gradually increased up to 95\% of the total surface area.

Experiment~2 involved an obstacle and 4 new objects to be placed on the table. The number of movable objects that were initially on the table was the control variable in this condition. The number of movable objects on the table  was increased from 4 to 36, such that the total footprint area covered by all objects and the obstacle were gradually increased up to 95\% of the total surface area.

Experiment~3 involved 4 initial objects randomly placed on the surface in collision-free configurations and  4 new objects to be added to the surface. The number of obstacles on the table was the control variable in this condition. The number of obstacles was increased from 4 to 32, such that the total footprint area covered by all objects and the obstacles were gradually increased up to 95\% of the total surface area.

Five algorithms were compared, where three of them correspond to the different levels of our nested local search:
\begin{itemize}
 \item innermost search (\emph{Inner}) based on potential fields,
 \item intermediate local search (\emph{Intermediate}) wrapped around the potential field, and
 \item the outermost local search (\emph{Outer}) wrapped around the intermediate local search.
\end{itemize}
 Remaining two algorithms are taken as baselines to demonstrate effectiveness of our approach with respect to naive implementations. In particular,
 \begin{itemize}
 \item to highlight the benefits of using the innermost search, we have tested \emph{Random Sample} approach, which randomly samples new configurations for objects with uniform distribution until a collision-free placement is found.
 \item To highlight the benefits of using our intermediate local search, we have tested \emph{Random Restart} which is implemented as the innermost search \emph{Inner} with random restarts.
 \end{itemize}

The performance of the algorithms were compared based on three metrics:
\begin{itemize}
\item efficiency, measured by the average CPU time spend to calculate a solution,
\item quality, measured by the average number of objects moved that were initially on the table, and
\item success rate, measured by percentage of trials that converge to a collision-free final configuration before a time-out is reached.
\end{itemize}
Efficiency and quality metrics are computed for partial solutions of unsuccessful trials that return configurations with collision(s).

Figure~\ref{fig:exp_plots} graphically summarizes the data collected from Experiments~1--3. In Figure~\ref{fig:exp_plots}, each row corresponds to an experiment, while columns present, CPU time, quality and success rate metrics, respectively.

We can observe the effect of increasing the number of new objects to be put on the table from the first row of Figure~\ref{fig:exp_plots}.
\begin{itemize}
\item In terms of average CPU time, \emph{Inner} outperforms \emph{Random Sample}, while \emph{Intermediate} consistently outperforms all other algorithms, including  \emph{Random Restart}. While \emph{Outer} is in general slower than \emph{Random Restart} for problems with surface coverage less than 75\%,  \emph{Outer} outperforms \emph{Random Restart} in highly cluttered environments, since it relies on \emph{Intermediate} when stuck at local minima.
\item In terms of solution quality, \emph{Inner} and \emph{Intermediate} perform quite similarly and move more objects as the table gets more cluttered, while \emph{Outer} performs significantly better than both. In particular, up to 30\% surface coverage, \emph{Outer} could find solutions that do not require any movements of the objects initially on the table. As the clutter increases, it becomes necessary to move some of these objects, but the number of moved objects stays significantly less than those in the solutions computed by \emph{Inner} and \emph{Intermediate}. It is important to note that the baseline algorithms, \emph{Random Sample} and \emph{Random Restart}, consistently result in low solution quality, even when compared to algorithms that do not distinguish between original and new objects, such as \emph{Intermediate}.
\item In terms of success rate, we note that \emph{Intermediate} and \emph{Outer} can solve almost all problems up to 85\%, and are the only algorithms capable of solving problems more cluttered than 85\%, although the success rate drops to about 40\%. While \emph{Random Sample} outperforms \emph{Inner} for simple problems where the surface coverage is less than 30\%, \emph{Inner} can solve some problems of up to 75\% surface coverage, while \emph{Random Sample} cannot solve any problems that has more than 40\% surface coverage.
\end{itemize}

We can observe the effect of increasing the number of original objects on the table from the second row of Figure~\ref{fig:exp_plots}. Note that these objects can be moved but are desired not to be relocated; hence, the quality of solutions becomes more emphasized in these experiments.
\begin{itemize}
\item In terms of average CPU time, once again, \emph{Inner} outperforms the \emph{Random Sample}, and \emph{Intermediate} outperforms \emph{Random Restart}. From these experiments, we can observe how increasing the number of original objects affects \emph{Outer}, causing it to have a sharp increase in computation time at about 50\% surface coverage.
\item The trend observed from the first set of experiments regarding solution quality can be seen again here. As \emph{Outer} is the only algorithm that attempts to improve solution quality, it consistently moves fewer original objects across all surface coverage levels.
\item The success rate of \emph{Outer} varies over surface coverage levels, but is always worse than \emph{Intermediate}. In particular, given that \emph{Outer} aims to solve a more constrained version of the problem solved by \emph{Intermediate}, it is not surprising that it fails more often. In this experiment set, \emph{Intermediate} is the only algorithm that is capable of solving a significant portion of problems with 95\% surface coverage.

\end{itemize}

We can observe the effect of increasing the number of obstacles on the table from the third row of Figure~\ref{fig:exp_plots}. Note that since obstacles cannot be moved, these instances prove to be much harder than the previous experiments, as the percentage of surface coverage increases.
\begin{itemize}
\item In terms of average CPU time,  \emph{Random Restart}, \emph{Intermediate}, and \emph{Outer} perform similarly up to 65\% clutter, after which \emph{Random Restart} baseline falls behind in computation time. The computation time for \emph{Outer} and \emph{Intermediate} display great similarity until about 85\%, after which no algorithm can solve even a single instance. Note that as the obstacles increase, the problem become highly constrained and objects initially located on the table are trapped; hence, object movements are much less in this experiment.
\item In terms of solution quality, \emph{Random Sample} and \emph{Random Restart} baselines perform quite similarly and move almost all objects initially on the table, while \emph{Outer} and \emph{Intermediate} perform slightly better than both baselines. \emph{Inner} performs the best in terms of solution quality, but this is only due to the constrained nature of the problem mentioned earlier.
\item \emph{Inner} and \emph{Intermediate} display significantly different trends in terms of success rate. In particular \emph{Inner} starts failing quite early and can only solve about 30\% of instances at 55\% clutter, while \emph{Intermediate} and \emph{Outer} consistently find over 80\% of the solutions up to 85\% clutter, after which all approaches fail.
\end{itemize}

The results of these experiments indicate the usefulness of all three nested local searches proposed by our method. \emph{Outer} significantly improves solution quality by minimizing movements of objects on the table, \emph{Intermediate} significantly improves success rate by allowing replacements when stuck in local optima, and \emph{Inner} significantly improves CPU time over \emph{Random Sampling} especially for cluttered scenarios.

\section{Solutions to Benchmarks Instances}

To demonstrate the ability of our local search approach to solve a large variety of placement problems, we have also tested it with several difficult benchmark scenarios. These benchmarks have been engineered to result in difficult instances, by introduction of  non-convex objects, confined surfaces, and very specific configurations that result in feasible placements.

\subsection{Confined Placement Benchmark}

\begin{figure}[ht]
    \centering
    \resizebox{\columnwidth}{!}{\includegraphics{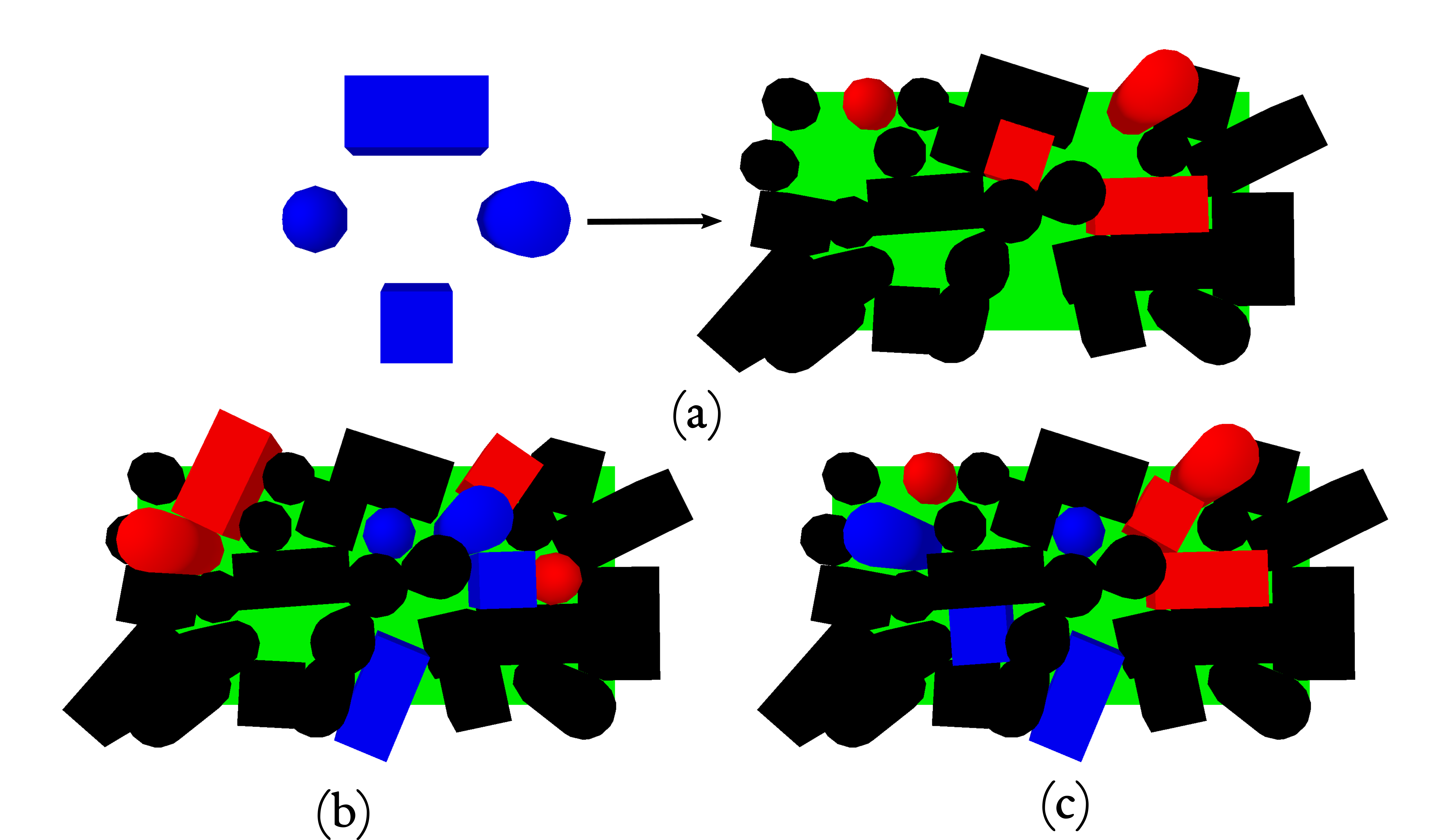}}
    \caption{Confined Placement Benchmark:  (a) describes the problem. (b) and (c) present the solutions computed using \textit{Intermediate} and \textit{Outer}, respectively.}
        \vspace{-\baselineskip}

    \label{fig:many_obs}
\end{figure}

In the confined placement benchmark, 4 objects needs to be placed on a surface highly cluttered with randomly placed of obstacles and 4 movable objects, as shown in Figure~\ref{fig:many_obs}(a). All objects and obstacles are selected as clones of the same 4 basic convex shapes.
This benchmark is challenging as it requires the placement algorithms to be able to handle highly restricted spaces. Figure~\ref{fig:many_obs}(b) and~\ref{fig:many_obs}(c) present solutions computed by  \textit{Intermediate} and  \textit{Outer}, respectively.
The solution computed by \textit{Outer} does requires only 1 of the 4 movable objects initially on the surface to be relocated, while the solution computed by \textit{Intermediate} relocates all of these objects.

\subsection{Tight Placement Benchmark}

\begin{figure}[ht]
    \centering
    \resizebox{1.05\columnwidth}{!}{\includegraphics{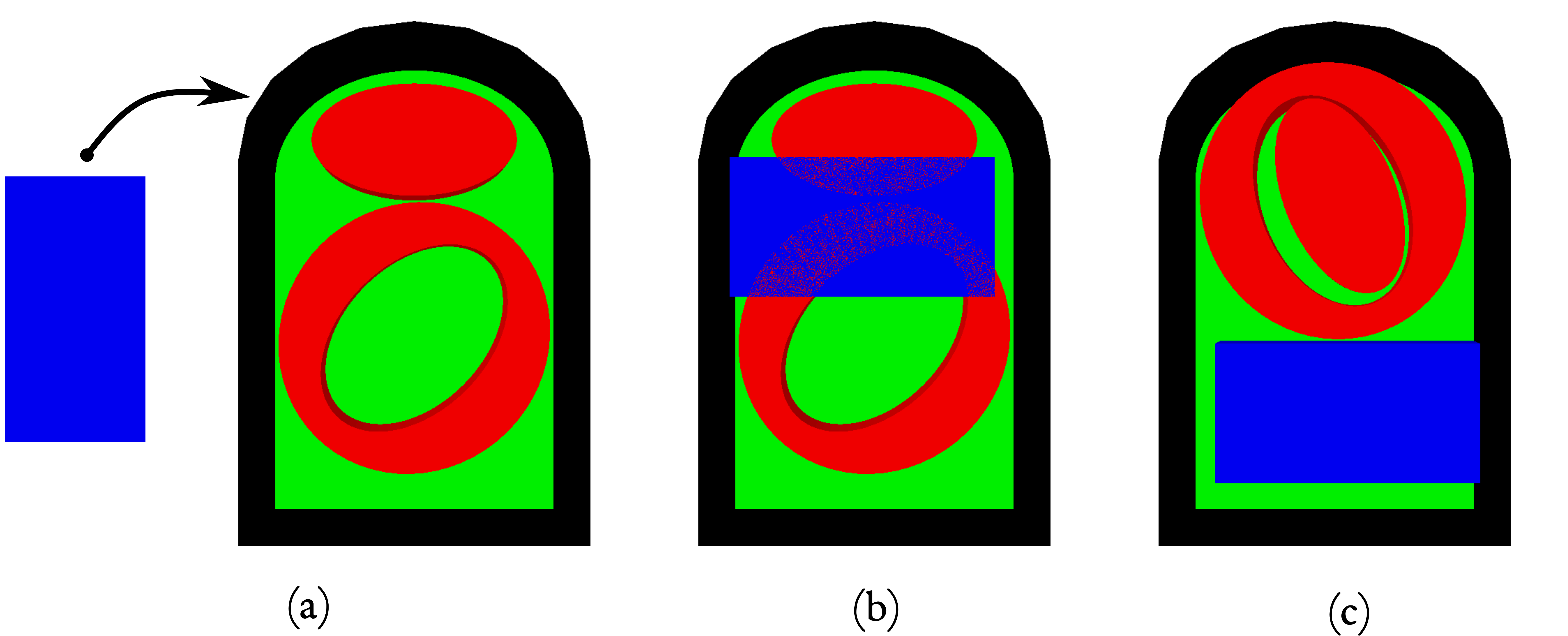}}
    \caption{Tight Placement Scenario. (a) describes the problem, (b) presents a random placement, and (c) shows the final solution computed by our algorithm. }
    \vspace{-\baselineskip}
    \label{fig:tight_placement}
\end{figure}

The tight placement benchmark  shown in Figure~\ref{fig:tight_placement}(a), where the goal is to find a placement for the blue box upon the green surface, has been introduced in~\cite{havur2014geometric}. This benchmark is challenging as it requires the objects on the surface to be rearranged to very specific configurations such that a solution can be computed. Furthermore, this benchmark includes non-convex objects with holes, which need to be utilized to compute a solution.  Figure~\ref{fig:tight_placement}(b) depicts a random  placement for this problem, demonstrating the need for rearrangement of objects on the surface.  Figure~\ref{fig:tight_placement}(c) presents the solution computed by our algorithm.

\begin{figure}[ht!]
    \centering
    \resizebox{.85\columnwidth}{!}{\includegraphics{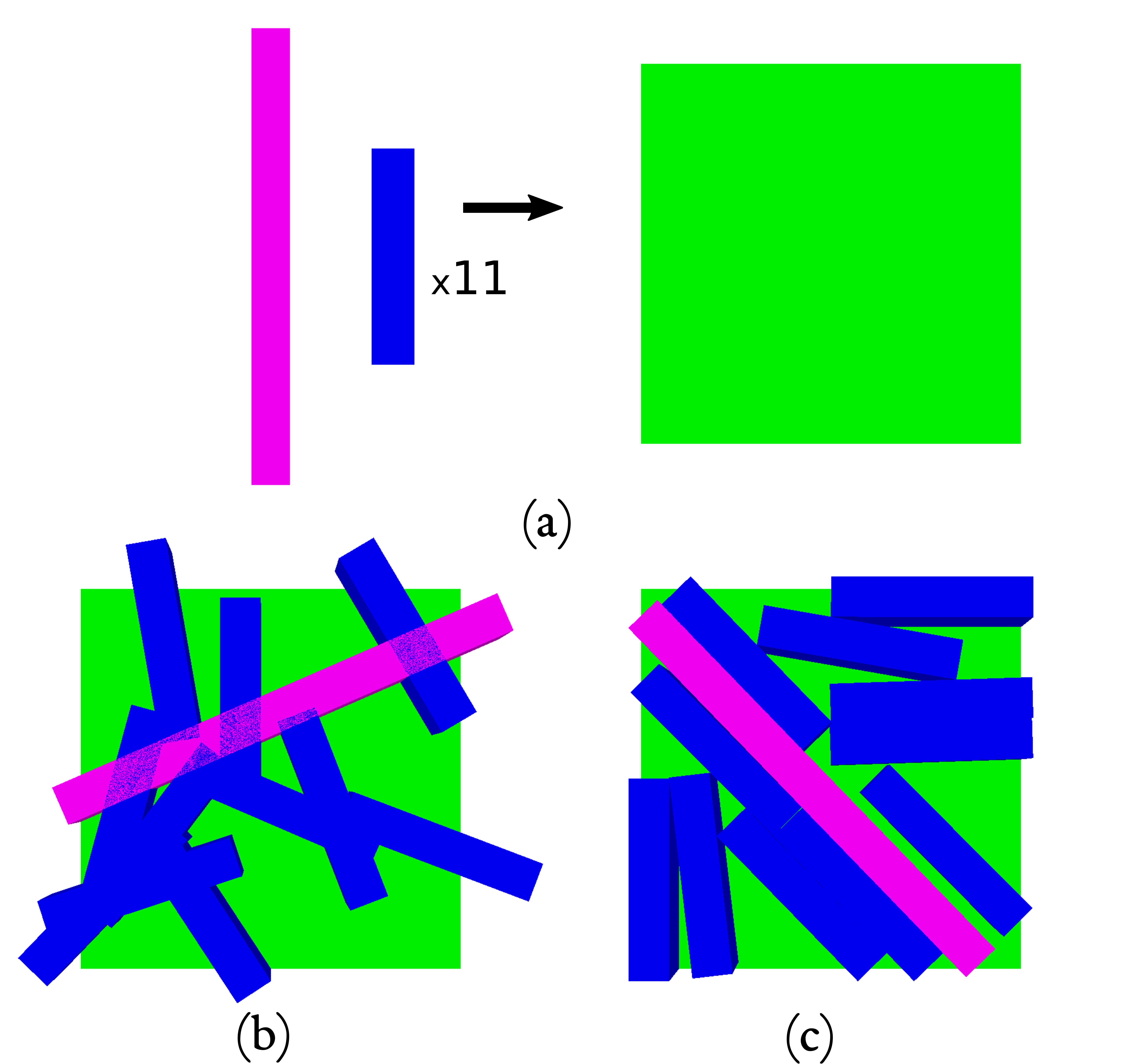}}
    \vspace{-.5\baselineskip}
    \caption{Elongated Object Benchmark.  (a) describes the problem, (b) presents a random placement demonstrating the unusable area introduced by the long object, and (c) shows the final solution computed by our algorithm.}
    \label{fig:elongated}
\end{figure}

\subsection{Elongated Object Benchmark}

\begin{figure}[ht]
    \centering
    \resizebox{0.85\columnwidth}{!}{\includegraphics{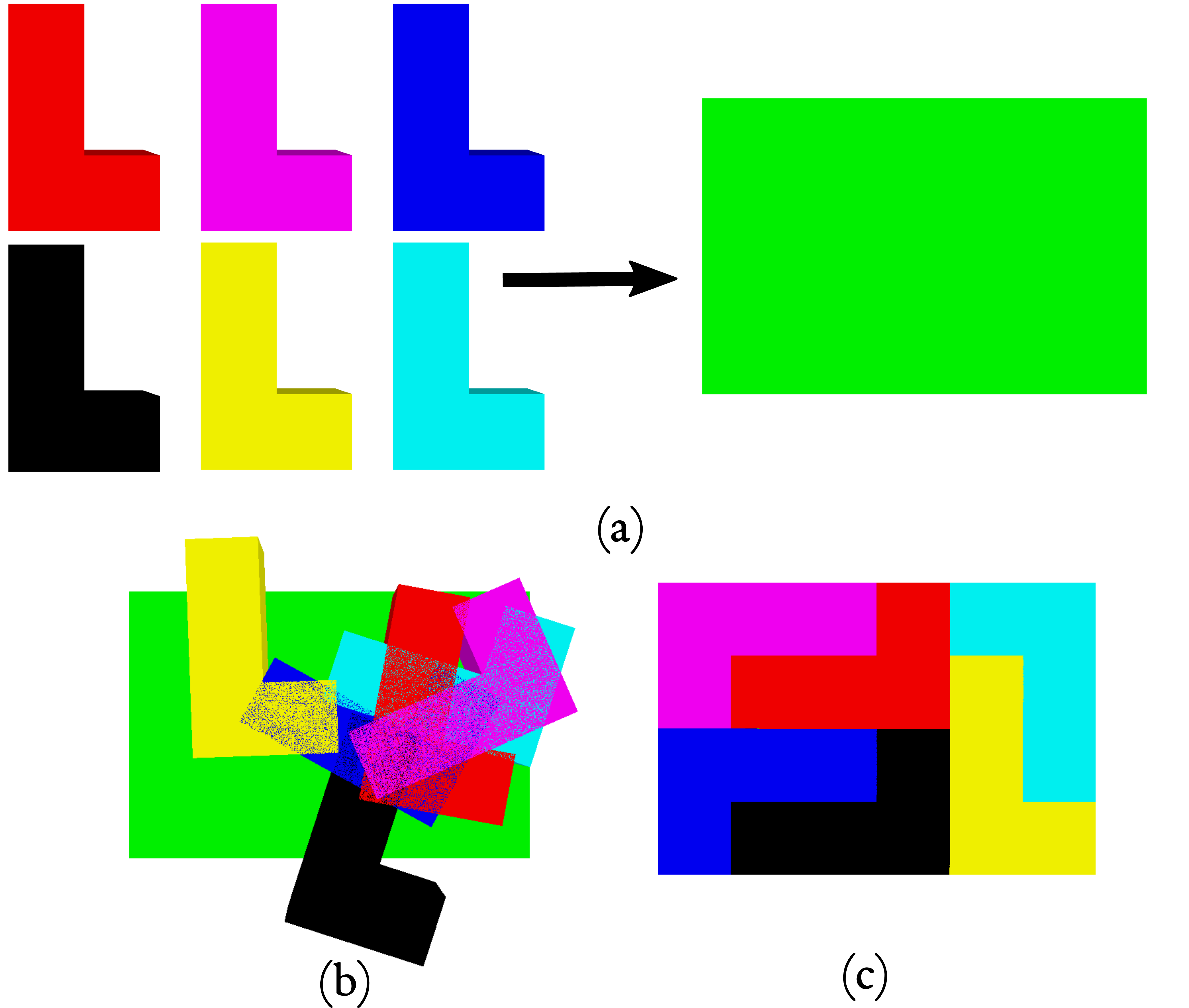}}
    \caption{2D L-Shape Placement Benchmark. (a) describes the problem, (b) presents a random placement demonstrating the need for perfect fitting of objects with one another in 2D, and (c) shows  the final solution computed by our algorithm.}

    \label{fig:lshapes}
\end{figure}

In the elongated object benchmark, the length of slender objects are designed to pose a challenge, as shown in Figure~\ref{fig:elongated}(a). In particular, one of the objects is selected to have a length greater than the width of the square surface, while 11 other slender objects are set to have a length that  is equal to the half the width of the square surface. This benchmark is difficult, as the placement of the longest object introduces unusable space on the square surface, rendering convergence to a solution significantly more difficult. Figure~\ref{fig:elongated}(b) depicts a random placement for this  problem, demonstrating the unusable area introduced by the long object.  Figure~\ref{fig:elongated}(c) presents the solution computed by our algorithm.

\subsection{L-Shape Placement Benchmarks}

\begin{figure}[ht]
    \centering
        \vspace{-0.5\baselineskip}

    \resizebox{0.85\columnwidth}{!}{\includegraphics{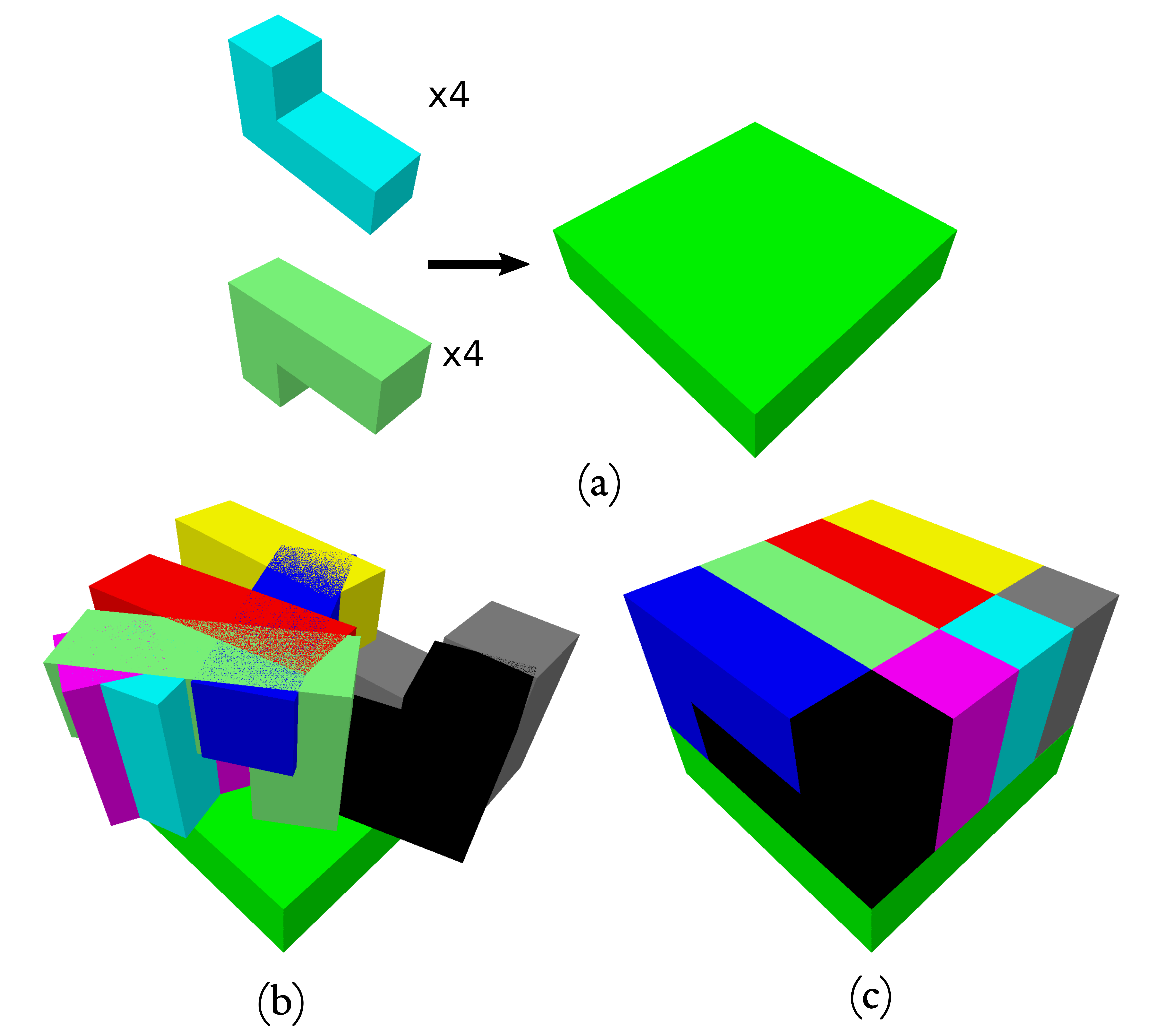}}
    \caption{3D L-Shape Placement Benchmark. (a) describes the problem, (b) presents a random placement demonstrating the need for perfect fitting of objects with one another in 3D, and (c) shows  the final solution computed by our algorithm.}
    \label{fig:lshapes-z}
    
\end{figure}

L-shape benchmarks are commonly utilized to demonstrate capabilities of packing algorithms~\cite{jain1998two}. L-shape placement benchmarks adapt these to placement problems with 2D and 3D variations, as shown in  Figures~\ref{fig:lshapes} and~\ref{fig:lshapes-z}, respectively.
In  L-shape placement benchmarks, the objects are required to be placed on a surface with no overhangs, and the total contact area of the L-shapes to be placed are equal to the area of the surface. In the 2D version, the L-shapes are configured such the objects have uniform height along the z-axis and are non-convex on parallel to the surface, while in the 3D version, the L-shapes are configured such the objects are uniform parallel to the surface  and are non-convex  along the z-axis.  Figures~\ref{fig:lshapes}(b) and~\ref{fig:lshapes-z}(b) present random placements for 2D and 3D benchmarks, demonstrating the difficulty of finding a collision free solution with no overhangs from the table. Figures~\ref{fig:lshapes}(c) and~\ref{fig:lshapes-z}(c)  present the solutions computed by our algorithm to the 2D and 3D benchmarks, respectively.

\section{CONCLUSION}

Given a surface cluttered with (unmovable) obstacles and movable objects, and a new set of objects, the object placement problem asks for a collision-free placement of all the objects on the surface. We have introduced a novel algorithm to solve this problem, utilizing nested local search algorithms with multi-objective optimizations: the innermost search tries to minimize the total penetration depths of objects, the intermediate local search further tries to minimize the number of collisions, and the outermost local search further tries to minimize the changes in object poses. Each level of the search is guided by heuristics: the innermost search utilizes a potential field method over a physics-based engine, the intermediate local search gradually utilizes re-placements of objects to avoid local minima, and the outermost local search gradually relaxes the constraints that specify how far objects can be displaced.  In that sense, our method introduces novel mathematical search models with nested multi-objective optimizations, and algorithms that further utilize heuristics to avoid local minima. On the other hand,  due to local searches, our algorithm is more about local optimization and is likely not to reach the global optima.

Comprehensive experimental evaluations demonstrate high computational efficiency and  success rate  of our method, as well as good quality of solutions. Furthermore, difficult benchmark scenarios that include non-convex objects, confined surfaces, and very specific configurations that result in feasible placements, can also be solved with our method.

\bibliographystyle{IEEEtran}
\bibliography{references}

\begin{thebibliography}{10}
\providecommand{\url}[1]{#1}
\csname url@samestyle\endcsname
\providecommand{\newblock}{\relax}
\providecommand{\bibinfo}[2]{#2}
\providecommand{\BIBentrySTDinterwordspacing}{\spaceskip=0pt\relax}
\providecommand{\BIBentryALTinterwordstretchfactor}{4}
\providecommand{\BIBentryALTinterwordspacing}{\spaceskip=\fontdimen2\font plus
\BIBentryALTinterwordstretchfactor\fontdimen3\font minus
  \fontdimen4\font\relax}
\providecommand{\BIBforeignlanguage}[2]{{%
\expandafter\ifx\csname l@#1\endcsname\relax
\typeout{** WARNING: IEEEtran.bst: No hyphenation pattern has been}%
\typeout{** loaded for the language `#1'. Using the pattern for}%
\typeout{** the default language instead.}%
\else
\language=\csname l@#1\endcsname
\fi
#2}}
\providecommand{\BIBdecl}{\relax}
\BIBdecl

\bibitem{stilman2005navigation}
M.~Stilman and J.~J. Kuffner, ``Navigation among movable obstacles: Real-time
  reasoning in complex environments,'' \emph{International Journal of Humanoid
  Robotics}, vol.~2, no.~04, pp. 479--503, 2005.

\bibitem{stilman2008planning}
M.~Stilman and J.~Kuffner, ``Planning among movable obstacles with artificial
  constraints,'' \emph{The International Journal of Robotics Research},
  vol.~27, no. 11-12, pp. 1295--1307, 2008.

\bibitem{wilfong1991motion}
G.~Wilfong, ``Motion planning in the presence of movable obstacles,''
  \emph{Annals of Mathematics and Artificial Intelligence}, vol.~3, no.~1, pp.
  131--150, 1991.

\bibitem{demaine2003pushing}
E.~D. Demaine, M.~L. Demaine, M.~Hoffmann, and J.~O'Rourke, ``Pushing blocks is
  hard,'' \emph{Computational Geometry}, vol.~26, no.~1, pp. 21--36, 2003.

\bibitem{okada2004environment}
K.~Okada, A.~Haneda, H.~Nakai, M.~Inaba, and H.~Inoue, ``Environment
  manipulation planner for humanoid robots using task graph that generates
  action sequence,'' in \emph{Intelligent Robots and Systems, 2004.(IROS 2004).
  Proceedings. 2004 IEEE/RSJ International Conference on}, vol.~2.\hskip 1em
  plus 0.5em minus 0.4em\relax IEEE, 2004, pp. 1174--1179.

\bibitem{stilman2007manipulation}
M.~Stilman, J.-U. Schamburek, J.~Kuffner, and T.~Asfour, ``Manipulation
  planning among movable obstacles.''\hskip 1em plus 0.5em minus 0.4em\relax
  Georgia Institute of Technology, 2007.

\bibitem{dogar2012planning}
M.~R. Dogar and S.~S. Srinivasa, ``A planning framework for non-prehensile
  manipulation under clutter and uncertainty,'' \emph{Autonomous Robots},
  vol.~33, no.~3, pp. 217--236, 2012.

\bibitem{barry2013manipulation}
J.~Barry, K.~Hsiao, L.~P. Kaelbling, and T.~Lozano-P{\'e}rez, ``Manipulation
  with multiple action types,'' in \emph{Experimental Robotics}.\hskip 1em plus
  0.5em minus 0.4em\relax Springer, 2013, pp. 531--545.

\bibitem{cosgun2011push}
A.~Cosgun, T.~Hermans, V.~Emeli, and M.~Stilman, ``Push planning for object
  placement on cluttered table surfaces,'' in \emph{Intelligent Robots and
  Systems (IROS), 2011 IEEE/RSJ International Conference on}.\hskip 1em plus
  0.5em minus 0.4em\relax IEEE, 2011, pp. 4627--4632.

\bibitem{havur2014geometric}
G.~Havur, G.~Ozbilgin, E.~Erdem, and V.~Patoglu, ``Geometric rearrangement of
  multiple movable objects on cluttered surfaces: A hybrid reasoning
  approach,'' in \emph{Robotics and Automation (ICRA), 2014 IEEE International
  Conference on}.\hskip 1em plus 0.5em minus 0.4em\relax IEEE, 2014, pp.
  445--452.

\bibitem{krontiris2014rearranging}
A.~Krontiris, R.~Shome, A.~Dobson, A.~Kimmel, and K.~Bekris, ``Rearranging
  similar objects with a manipulator using pebble graphs,'' in \emph{Humanoid
  Robots (Humanoids), 2014 14th IEEE-RAS International Conference on}.\hskip
  1em plus 0.5em minus 0.4em\relax IEEE, 2014, pp. 1081--1087.

\bibitem{krontiris2015dealing}
A.~Krontiris and K.~E. Bekris, ``Dealing with difficult instances of object
  rearrangement.'' in \emph{Robotics: Science and Systems}, 2015.

\bibitem{krontiris2016efficiently}
------, ``Efficiently solving general rearrangement tasks: A fast extension
  primitive for an incremental sampling-based planner,'' in \emph{Robotics and
  Automation (ICRA), 2016 IEEE International Conference on}.\hskip 1em plus
  0.5em minus 0.4em\relax IEEE, 2016, pp. 3924--3931.

\bibitem{han2017high}
S.~D. Han, N.~M. Stiffler, A.~Krontiris, K.~E. Bekris, and J.~Yu,
  ``High-quality tabletop rearrangement with overhand grasps: Hardness results
  and fast methods,'' in \emph{Proceedings of Robotics: Science and Systems},
  2017.

\bibitem{kang2018automated}
M.~Kang, Y.~Kwon, and S.-E. Yoon, ``Automated task planning using object
  arrangement optimization,'' in \emph{2018 15th International Conference on
  Ubiquitous Robots (UR)}.\hskip 1em plus 0.5em minus 0.4em\relax IEEE, 2018,
  pp. 334--341.

\bibitem{yu2011make}
L.~F. Yu, S.~K. Yeung, C.~K. Tang, D.~Terzopoulos, T.~F. Chan, and S.~J. Osher,
  ``Make it home: automatic optimization of furniture arrangement,'' 2011.

\bibitem{jiang2012learning}
Y.~Jiang, M.~Lim, C.~Zheng, and A.~Saxena, ``Learning to place new objects in a
  scene,'' \emph{The International Journal of Robotics Research}, vol.~31,
  no.~9, pp. 1021--1043, 2012.

\bibitem{jiang2012humanlearning}
Y.~Jiang, M.~Lim, and A.~Saxena, ``Learning object arrangements in 3d scenes
  using human context,'' in \emph{Proceedings of the 29th International
  Coference on International Conference on Machine Learning}.\hskip 1em plus
  0.5em minus 0.4em\relax Omnipress, 2012, pp. 907--914.

\bibitem{jiang2013hallucinating}
Y.~Jiang and A.~Saxena, ``Hallucinating humans for learning robotic placement
  of objects,'' in \emph{Experimental Robotics}.\hskip 1em plus 0.5em minus
  0.4em\relax Springer, 2013, pp. 921--937.

\bibitem{Chazelle1989}
B.~Chazelle, H.~Edelsbrunner, and L.~J. Guibas, ``The complexity of cutting
  complexes,'' \emph{Discrete {\&} Computational Geometry}, vol.~4, no.~2, pp.
  139--181, 1989.

\bibitem{DYCKHOFF1990}
H.~Dyckhoff, ``A typology of cutting and packing problems,'' \emph{European
  Journal of Operational Research}, vol.~44, no.~2, pp. 145 -- 159, 1990.

\bibitem{Liu2015}
X.~Liu, J.-m. Liu, A.-x. Cao, and Z.-l. Yao, ``Hape3d---a new constructive
  algorithm for the 3d irregular packing problem,'' \emph{Frontiers of
  Information Technology {\&} Electronic Engineering}, vol.~16, no.~5, pp.
  380--390, 2015.

\bibitem{ROMANOVA2018}
T.~Romanova, J.~Bennell, Y.~Stoyan, and A.~Pankratov, ``Packing of concave
  polyhedra with continuous rotations using nonlinear optimisation,''
  \emph{European Journal of Operational Research}, vol. 268, no.~1, pp. 37 --
  53, 2018.

\bibitem{Ma2018}
Y.~Ma, Z.~Chen, W.~Hu, and W.~Wang, ``Packing irregular objects in 3d space via
  hybrid optimization,'' \emph{Computer Graphics Forum}, vol.~37, no.~5, pp.
  49--59, 2018.

\bibitem{EGEBLAD2009}
J.~Egeblad, B.~K. Nielsen, and M.~Brazil, ``Translational packing of arbitrary
  polytopes,'' \emph{Computational Geometry}, vol.~42, no.~4, pp. 269 -- 288,
  2009.

\bibitem{martello2000}
S.~Martello, D.~Pisinger, and D.~Vigo, ``The three-dimensional bin packing
  problem,'' \emph{Operations Research}, vol.~48, no.~2, pp. 256--267, 2000.

\bibitem{Fasano2013}
G.~Fasano, ``A global optimization point of view to handle non-standard object
  packing problems,'' \emph{Journal of Global Optimization}, vol.~55, no.~2,
  pp. 279--299, 2013.

\bibitem{Khatib1086}
O.~Khatib, ``Real-time obstacle avoidance for manipulators and mobile robots,''
  \emph{The International Journal of Robotics Research}, vol.~5, no.~1, pp.
  90--98, 1986.

\bibitem{Patoglu2005a}
V.~Patoglu and R.~B. Gillespie, ``A feedback stabilized minimum distance
  maintenance for convex parametric surfaces,'' \emph{IEEE Transactions on
  Robotics}, vol.~25, no.~5, 2005.

\bibitem{coumans2010bullet}
E.~Coumans, ``Bullet physics engine,'' \emph{Open Source Software:
  http://bulletphysics. org}, vol.~1, 2010.

\bibitem{jain1998two}
S.~Jain and H.~C. Gea, ``Two-dimensional packing problems using genetic
  algorithms,'' \emph{Engineering with Computers}, vol.~14, no.~3, pp.
  206--213, 1998.

\end{thebibliography}

\end{document}